\documentclass[11pt]{article}

\usepackage[preprint]{acl}

\usepackage{times}
\usepackage{amsmath}
\usepackage{amssymb}
\usepackage{latexsym}
\usepackage{booktabs}
\usepackage{multirow}
\usepackage{enumitem}
\usepackage{pifont}
\usepackage[table]{xcolor}
\usepackage{arydshln}
\usepackage{float}
\usepackage[most]{tcolorbox}
\usepackage{xcolor}
\usepackage{listings}
\definecolor{lightblue}{RGB}{230, 245, 255}
\definecolor{topone}{RGB}{255, 226, 190}
\definecolor{toptwo}{RGB}{255, 239, 220}
\definecolor{topthree}{RGB}{255, 248, 238}
\definecolor{myblue}{RGB}{220,235,247}
\newcommand{\first}[1]{\cellcolor{topone}\textbf{#1}}
\newcommand{\second}[1]{\cellcolor{toptwo}#1}
\newcommand{\third}[1]{\cellcolor{topthree}#1}
\newcommand{\reddown}{\textcolor{red}{\scriptsize$\downarrow$}}
\newcommand{\greenup}{\textcolor{green!60!black}{\scriptsize$\uparrow$}}

\usepackage[T1]{fontenc}
\usepackage[utf8]{inputenc}

\usepackage{microtype}

\usepackage{inconsolata}

\usepackage{graphicx}

\title{\textit{READ More than What You See}:
Reinforcement Learning for Accurate and Coherent Audio Description Generations}

\author{
 \textbf{Bo Fang\textsuperscript{1}} \qquad 
 \textbf{Xinyao Zhang\textsuperscript{2,3}} \qquad 
 \textbf{Yuxin Song\textsuperscript{2}} \qquad 
 \textbf{Hui Zhang\textsuperscript{2}} \qquad 
 \\
 \textbf{Hang Zhou\textsuperscript{2}} \qquad 
 \textbf{Antoni B. Chan\textsuperscript{1}}
\\
\\
 \textsuperscript{1}City University of Hong Kong,
 \textsuperscript{2}Baidu Inc,
 \textsuperscript{3}Tsinghua University
\\
 \small{
   \textbf{Correspondence:} \href{bofang6-c@my.cityu.edu.hk}{bofang6-c@my.cityu.edu.hk}
 }
}

\begin{document}
\maketitle
\begin{abstract}

Audio Description aims to generate concise narrations of essential visual content in audiovisual media for blind and low-vision audiences. 
Existing methods either rely on prompting off-the-shelf multimodal models, which often mismatch AD style, or partially optimize training-based systems with next-token prediction, which under-explores model capacity and biases generation toward generic expressions. 
We present READ, the first reinforcement-learning (RL) framework for training-based AD generation. 
READ formulates AD as sequence-level optimization with reference-matching, length, and format rewards, and further introduces a dedicated coherence reward under context-aware supervision to promote narratively coherent descriptions. 
Experiments on MAD-Eval, CMD-AD, and TV-AD show that READ substantially outperforms prior 
% training-free and training-based 
methods across diverse evaluation metrics. Our results highlight RL as a promising paradigm for accurate and coherent AD generation.
Our codes, models, and benchmark results will be publicly available.

\end{abstract}

% =============================================================================
\section{Introduction}
\label{sec:intro}

% what is AD and its application
\textit{Audio Descriptions} (ADs) refer to the generation of concise natural-language narrations that convey essential visual content in audiovisual media for blind and low-vision audiences. 
Unlike dense video captioning~\citep{krishna2017dense}, ADs are tightly constrained by narrative contexts and delivery time, as descriptions must be inserted around dialogues without disrupting the original audio~\citep{pavel2020rescribe}. 
High-quality AD therefore requires both \textbf{accurate} selection of salient plot-relevant visual information such as actions, expressions, and scene changes, and \textbf{coherent} narration that integrates naturally with contexts and story understanding. 
%
% This makes AD important not only for media accessibility, but also for inclusive participation in audiovisual storytelling.

% current research development & their limitations
% Recent advances in powerful vision-language models (VLMs), such as the QwenVL series~\citep{wang2024qwen2vl,bai2025qwen3vl}, as well as commercial multimodal systems like GPT-4o~\cite{hurst2024gpt4o,singh2025gpt5}, have substantially accelerated research on AD. These developments have spurred growing interest in both \textit{training-free} and \textit{training-based} AD paradigms.

Recent advances in large vision-language models (VLMs)~\citep{wang2024qwen2vl,bai2025qwen3vl,hurst2024gpt4o} have rapidly pushed forward automatic AD generation, giving rise to two main paradigms: \textit{training-free} prompting and \textit{training-based} models. 
The former rely on off-the-shelf multimodal models with carefully designed prompts and contextual cues~\cite{zhang2024mm,xie2024autoad0,xie2025shot}, while the latter adapt models to AD domain, typically by learning a projector from CLIP~\citep{radford2021CLIP} visual features into the language model space~\citep{han2023autoad,han2024autoad3,fang2025distinctad,wang2025uniad}.

% Training-free AD methods typically exploit off-the-shelf multimodal models through carefully designed visual prompts and auxiliary contextual signals, such as in-context clips~\citep{zhang2024mm}, center-character cues~\cite{xie2024autoad0}, and shot-level grounding~\citep{xie2025shot}. They often further employ LLM-based rephrasing or summarization to produce descriptions that more closely resemble human-authored references. 
% %
% In parallel, end-to-end training-based approaches usually learn a projector that maps CLIP-based~\cite{radford2021CLIP} visual features into the language model space for AD generation. 

However, current methods still face clear limitations. 
\textit{Training-free} approaches often remain mismatched to the style of ADs, and their performance is inherently limited without task-specific learning. 
\textit{Training-based} methods, on the other hand, can generate more in-domain descriptions, but most simply learn a lightweight projector on top of frozen modules, leaving the underlying model capacity for AD \textbf{in}sufficiently exploited. 
More importantly, prior end-to-end methods are typically optimized solely with next-token prediction loss, which encourages imitation of frequent patterns in the training data and can bias the model toward generic expressions rather than truly informative descriptions. 
While such objectives may improve local reference matching, they do not directly optimize sequence-level AD properties, particularly accurate content specification and consistency with the surrounding contexts.

% our methods: two dimensions - accuracy and coherence
% We propose \textbf{READ}, a \textbf{RE}inforcement learning framework that targets accuracy and narrative coherence simultaneously for \textbf{A}udio \textbf{D}escription generation.
% % motivation
% Our strong motivation comes from recent success in recent reasoning models using reinforcement learning~\citep{shao2024deepseekmath,guo2025deepseekr1}, where has greatly facilitated open-ended visual understanding with reliable rewards~\citep{feng2025onethinker}.

Our motivation comes from recent progress in RL-based post-training,
which has strongly improved reasoning in LLMs beyond next-token prediction~\citep{shao2024deepseekmath,guo2025deepseekr1}, and is increasingly being extended to multimodal visual understanding~\citep{huang2025visior1,deng2025openvlthinker,feng2025onethinker}. 
We view audio description in a similar light: \textit{generating ADs is not merely a token-level matching problem, but an open-ended generation task that involves non-trivial reasoning over which visual content is salient, how it should be verbalized in AD style, and how each description should remain compatible with the contexts}. 
This motivates us to go beyond conventional supervised fine-tuning and investigate RL for AD generation.

% accuracy: core driven reward
We propose \textbf{READ}, a \textbf{RE}inforcement-learning based framework for automatic \textbf{A}udio \textbf{D}escription generation. 
% 进入的不太顺畅，需要简短补充下GRPO
% \textcolor{red}{Specifically,
% READ uses a sequence-level reward based on the ROUGE~\citep{lin2004ROUGE} score between sampled rollouts and human-authored AD references, providing a simple yet effective supervision signal for this open-ended generation task. }
At its core, READ treats AD as a sequence level optimization problem, rather than relying solely on token-level supervision.
Accordingly, the baseline framework is driven by a reference-matching reward that encourages generated descriptions to remain faithful to human-authored ADs at the sequence level.
In practice, we instantiate this objective with a ROUGE-based reward computed between sampled rollouts and reference ADs, which provides a simple yet effective learning signal for this open-ended generation task.
% %
To account for the temporal constraints of AD, we further introduce a length reward that encourages clip-appropriate word counts, so that generated descriptions can be delivered within the available narration window. 
Additionally, we add a format reward to regularize the output structure, requiring intermediate reasoning and final descriptions to be enclosed in \texttt{<think>} and \texttt{<answer>} tags, respectively.
Together, these designs establish a strong RL baseline for AD.

% coherence design
To further promote narrative coherence, we extend READ beyond clip-level supervision by incorporating additional training instances of the form \texttt{\{\textit{context},\textit{clip}\}$\rightarrow$\textit{AD}}, so that the model is encouraged to generate descriptions that are compatible with the surrounding ADs. 
Building on this formulation, we introduce a dedicated coherence reward that explicitly favors contextually coherent generations. 
Concretely, this reward is based on a lightweight AD-adapted LLM, which evaluates the conditional plausibility of a generated description given its context. 
To prevent trivial gains from simply repeating contextual words, the coherence signal is computed in a way that discounts overlapping n-grams with the context. 
Our coherence reward is applied only when the generation already achieves reasonable content fidelity, allowing READ to improve narrative consistency without sacrificing descriptive accuracy.
Extensive experiments show that sequence-level RL enables READ to substantially surpass prior methods on conventional captioning metrics, while also yielding significant improvements on AD-specific evaluation measures.

% contributions
To summarize, our contributions are three-fold:
\begin{itemize}[leftmargin=0pt]
\item 
% To the best of our knowledge, READ is the first work to investigate reinforcement learning for automatic audio description generation, providing a strong foundation for training-based AD models.
To the best of our knowledge, READ is the first training-based AD framework to incorporate reinforcement learning, establishing a new optimization paradigm for automatic AD generation.

\item 
% In addition to improving fidelity to human-authored references, READ explicitly optimizes for narrative coherence, encouraging context-aware descriptions that better fit the surrounding story.
Beyond improving fidelity to human-authored references, READ further introduces a dedicated coherence reward into RL training to promote narratively coherent, context-aware ADs.

\item 
Extensive evaluations on MAD-Eval~\citep{soldan2022mad}, CMD-AD~\citep{han2024autoad3}, and TV-AD~\cite{xie2024autoad0} show that READ substantially and consistently outperforms prior training-free and training-based methods, achieving large gains across a wide range of evaluation metrics.
\end{itemize}

% =============================================================================
\section{Related Work}
\label{sec:related}

\noindent\textbf{AD generation.}
Early AD systems mainly relied on specialized authoring tools~\citep{branje2012livedescribe} and skilled human contributors.
Increasing efforts have been made to AD research with the rapid development of powerful LLMs~\citep{radford2019gpt-2,touvron2023llama2,grattafiori2024llama3} and VLMs~\citep{radford2021CLIP,wang2024qwen2vl,hurst2024gpt4o}.

For automatic AD generation, training-free approaches~\citep{chu2024llm-ad,ye2024mmad,zhang2024mm,xie2024autoad0,xie2025shot} typically customize their AD outputs by prompting foundation models together with handcrafted pipelines and manually designed visual cues (characters, contexts, shots, \textit{etc}). 
Yet, these methods still lag in performance due to noticeable domain discrepancy between general-purpose LLM corpora and ADs.
Alternatively, training-based methods~\citep{han2023autoad,han2023autoad2,han2024autoad3,lin2024movieseq,fang2025distinctad,wang2025uniad} usually adapt VLMs to AD by fine-tuning a lightweight connector that projects visual features into the LLM space and optimizing with next-token prediction. 
Despite better performance, their training objective still tends to favor generic and less informative descriptions.
In contrast, our READ introduces RL into automatic AD generation, treating AD as an open-ended generation task that requires more effective optimization beyond standard next-token supervision.

\noindent\textbf{RL for (vision-)language models.}
RL has been widely used for sequence-level optimization in tasks such as machine translation~\citep{ranzato2015sequence}, summarization~\citep{paulus2017deep}, and image captioning~\citep{rennie2017selfcritical}. 
More recently, RL post-training has played an increasingly important role in LLMs~\cite{ouyang2022instructGPT,guo2025deepseekr1} and multimodal models~\citep{huang2025visior1,deng2025openvlthinker}, substantially enhancing alignment, reasoning, and visual understanding.
However, such advances have not been systematically explored for automatic AD. 
READ fills this gap by introducing GRPO-based RL into training-based AD generation.

\noindent\textbf{Context and coherence in AD.}
Context-aware generation has long been studied in AD, where neighboring descriptions, shot-level cues, or broader video contexts are used to improve narrative continuity and distinctiveness~\citep{han2023autoad,lin2024movieseq,xie2025shot,khandelwal2025coherentAD}.
UniAD~\citep{wang2025uniad} and DistinctAD~\citep{fang2025distinctad} fine-tune VLMs over multiple consecutive clips to incorporate boarder temporal context.
However, most existing works rarely optimizing contextual coherence as a training objective explicitly, which results superficial coherence by repeating contextual content. 
In contrast, READ introduces an explicit coherence reward for AD generation and combines it with anti-copy masking to encourage context-compatible yet distinctive descriptions.

% =============================================================================
\section{Method}
\label{sec:method}

% This section first reviews GRPO in \S\ref{subsec:method_preliminary}, then presents the READ baseline in \S\ref{subsec:method_rlad}, and extends it with coherence RL in \S\ref{subsec:method_coherence}. The overall training objective is summarized in \S\ref{subsec:method_overll}.

\subsection{Preliminary}
\label{subsec:method_preliminary}
Recent advances in reinforcement-learning-based post-training have demonstrated strong effectiveness in improving complex reasoning behaviors in large language and multimodal models. 
In this work, we focus on the RL component and build our method on Group Relative Policy Optimization (GRPO)~\citep{shao2024deepseekmath}, a variant of Proximal Policy Optimization (PPO)~\citep{schulman2017proximal} that avoids training an additional value model and is thus more computationally efficient.
For each input $q$, GRPO samples a group of responses \{$o_1, o_2, \cdots, o_n$\} from the current policy model $\pi_{\theta}$. A reward function assigns scalar rewards \{$R_1, R_2, \cdots, R_n$\} to these responses, and the relative advantage of each response is estimated by normalizing its reward within the group:
\begin{align}
    A_i = \frac{R_i-\mathrm{mean}(\{R_1, R_2, \cdots, R_G\})}{\mathrm{std}(\{R_1, R_2, \cdots, R_n\})}.
\end{align}
This group-relative formulation encourages the policy to prefer responses with higher relative quality within each sampled group. 
GRPO then optimizes the policy using a clipped objective together with KL regularization toward a reference model, which stabilizes the training updates while preserving useful prior behaviors:
\begin{equation}
\begin{aligned}
    \mathcal{J}_{\mathrm{GRPO}} =& \mathbb{E}_{q,\{o_i\}}
    \Big[
        \tfrac{1}{G}\sum\nolimits_{i=1}^{G}
        \big( \min\big(\tfrac{\pi_{\theta}(o_i|q)}{\pi_{\theta_{\mathrm{old}}}(o_i|q)}A_i, \\
         & \mathrm{clip}(\tfrac{\pi_{\theta}(o_i|q)}{\pi_{\theta_{\mathrm{old}}}(o_i|q)}, 1-\epsilon, 1+\epsilon)A_i \big)- \\
         & \beta \mathbb{D}_{\mathrm{KL}}(\pi_{\theta} \parallel \pi_{\mathrm{ref}}) \big) 
    \Big], 
\end{aligned}
\label{eq:grpo}
\end{equation}
where $G$ is the rollout group size, $\epsilon$ is the clipping threshold, $\pi_{\mathrm{ref}}$ is the reference policy, and $\pi_{\theta_{\mathrm{old}}}$ denotes the old policy.
% used for sampling and policy-ratio computation.

\subsection{GRPO for AD Generation: A Baseline}
\label{subsec:method_rlad}

\begin{figure*}[t]
  \includegraphics[width=1.0\linewidth]{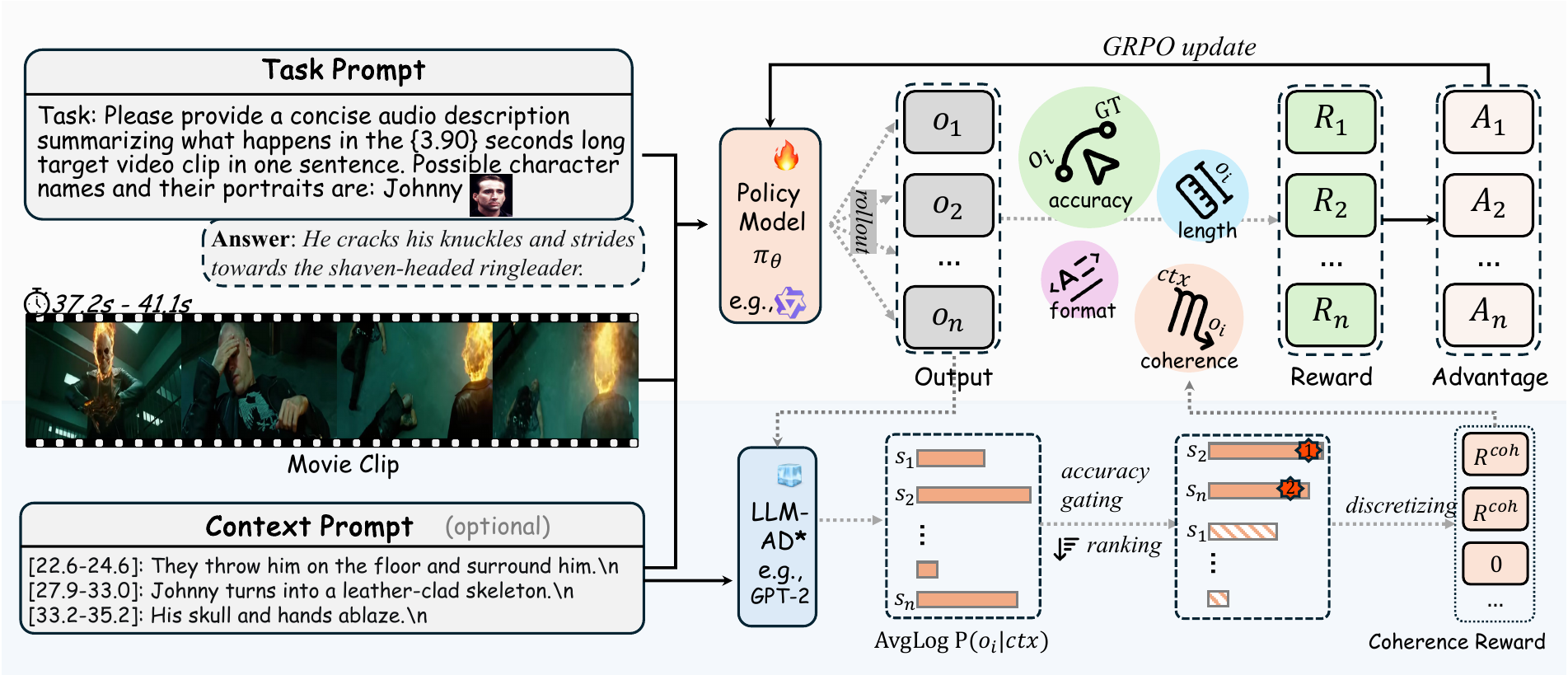}
  \vspace{-8mm}
  \caption{\textbf{Framework of READ.}
  For automatic AD generation, READ samples multiple candidates and optimizes them with GRPO using accuracy, format, and length rewards. 
  It further introduces a context-aware coherence reward to encourage context-compatible and narratively coherent AD generation.
  }
  \vspace{-4mm}
  \label{fig:method}
\end{figure*}

Building on the GRPO framework introduced above, we first construct a RL baseline for automatic AD generation, which we refer to as READ-base (Fig.~\ref{fig:method}). 
The baseline is designed to capture three basic requirements: 1) \textit{content fidelity to human-authored references}, 2) \textit{structural compliance with the desired output format}, and 3) \textit{temporal suitability for narration}. 
Accordingly, READ adopts three fundamental rewards, namely \emph{accuracy}, \emph{format}, and \emph{length}. We describe them below.

\paragraph{Accuracy reward.} The core reward in READ is designed to measure how well a sampled description matches the human-authored reference. 
Since AD generation is an open-ended language generation task, exact matching is overly restrictive and cannot adequately reflect partial correctness in content selection and phrasing. Following~\citep{feng2025onethinker}, we adopt a sequence-level reference reward based on the average of ROUGE score~\citep{lin2004rouge}.
Given a sampled response $o_i$ and its corresponding GroundTruth AD $g$, the accuracy reward is computed as:
\begin{equation}
R_i^{\mathrm{acc}}
=
\tfrac{1}{3}
\bigl(
\rho_1(o_i, g) + \rho_2(o_i, g) + \rho_L(o_i, g)
\bigr),
\label{eq:ROUGE}
\end{equation}
where $\rho_1$, $\rho_2$, and $\rho_L$ denote ROUGE-1, ROUGE-2, and ROUGE-L, respectively.
This reward provides an effective and efficient supervision signal for encouraging descriptions that remain faithful to the salient visual content of the target clip.

\paragraph{Format reward.}
% To regularize the output structure during RL, 
We impose a simple formatting constraint on model responses. 
Specifically, the model is required to place its intermediate content within \texttt{<think>...</think>} tags and the final AD output within \texttt{<answer>...</answer>} tags. 
This structured format standardizes model outputs and enables reliable extraction of the final AD description for reward computation.
We therefore assign a positive reward $R_i^{\mathrm{fmt}}$ to a sampled response if it satisfies the required tagging structure.

\paragraph{Length Reward.}
Since movie clips vary substantially in duration, the appropriate length of the corresponding AD also differs across samples. 
Descriptions that are too short may omit important plot details, while overly long descriptions may exceed the available narration window and thus be difficult to deliver in time.
We therefore introduce a length reward to encourage temporally appropriate AD generation.
Concretely, we use the word count of the GroundTruth AD $g$ as a simple proxy for the desired description length, and compare it with the word count of each sampled response. If the absolute length difference is within a predefined range, \textit{\textit{i.e.}}, $\left| \operatorname{wc}(o_i) - \operatorname{wc}(g) \right| \le \Lambda$, the response is regarded as length-compatible and receives a positive reward $R_i^{\mathrm{len}}$. (More analysis are in \S\ref{app:ad_len_sta}.)
% \begin{equation}
% R_i^{\mathrm{len}} =
% \begin{cases}
% 1, & \text{if } \left| \operatorname{wc}(o_i) - \operatorname{wc}(y) \right| \le \tau,\\
% 0, & \text{otherwise}.
% \end{cases}
% \end{equation}

\subsection{Context-Aware Coherence Reinforcement}
\label{subsec:method_coherence}
% While the READ baseline provides strong clip-level supervision, it treats each clip--AD pair in isolation. However, ADs are consumed as a narrative sequence, where a locally accurate description may still be incoherent if it ignores preceding events, shifts referents inconsistently, or repeats nearby context. Since reference-based rewards mainly capture similarity to the current GT AD, they provide limited supervision for contextual compatibility. We therefore introduce context-aware coherence reinforcement to encourage descriptions that are faithful to the current clip and coherent with surrounding ADs.
Although the READ-base improves clip-level fidelity, it optimizes each clip independently, leaving locally accurate outputs that may still be inconsistent with or redundant to surrounding ADs. 
To address this limitation, we introduce context-aware coherence reinforcement, which encourages each generated AD to remain faithful to the current clip while fitting the surrounding narrative context.
% Despite providing a strong RL baseline for AD, READ lacks explicit coherence modeling with surrounding contexts.
% To this end, we further extend READ with a context-aware coherence RL mechanism. 
%
Specifically, beyond original clip--AD pairs, we construct and merge context-aware training instances of the form $([c_1; \cdots; c_m], q) \rightarrow g$, where $[c_1, \ldots, c_m]$ are the temporal nearest contextual ADs associated with the current clip $q$. As shown in Fig.~\ref{fig:method}, the coherence reward is built as follows.

Since narrative coherence is meaningful only when a response is already reasonably faithful to the current clip, a sampled response is considered for coherence evaluation only if its ROUGE score (accuracy) exceeds a predefined threshold $\tau_{\mathrm{acc}}$. 
This \textbf{\textit{gating}} strategy prevents the model from receiving extra reward for contextually plausible but visually inaccurate descriptions.

% Next, for each eligible response $o_i$, we concatenate its contextual ADs $[c_1; \cdots; c_m]$ with $o_i$ and feed the sequence into LLM-AD$^{*}$, a lightweight language model fine-tuned on AD corpora\footnote{https://audiovault.net}, to estimate the average conditional log-probability of the response under the given context. 
% Formally, the coherence score of $o_i$ is computed as

\noindent\textbf{Coherence score.}
For each eligible response $o_i$, we concatenate its contextual ADs into a single context sequence $c_{1:m} = [c_1; \cdots; c_m]$, and feed $[c_{1:m}; o_i]$ into LLM-AD$^{*}$
% , a lightweight language model fine-tuned on AD corpora\footnote{https://audiovault.net}, 
to estimate the conditional likelihood of the response under the given context. 
LLM-AD$^{*}$ is a lightweight language model (\textit{e.g.}, GPT-2) fine-tuned on AD corpora, making it better aligned with the concise narration style, temporal continuity commonly observed in ADs. (See \S\ref{app:train_detail} for more details.)
We use it as a domain-adapted scorer to rank rollouts within the same group, rather than as an absolute evaluator of AD quality.

To avoid trivial gains from simply copying contextual content, we compute the coherence $s_i$ using a masked average log-probability, where 3-gram or 4-gram spans of response tokens are excluded if they overlap with the context sequence.
% response tokens belonging to context-overlapping $3$-gram or $4$-gram spans are excluded from scoring. 
%
Formally, the coherence score of $o_i$ is computed as
\begin{equation}
s_i
=
\frac{
\sum_{t=1}^{L_i}
m_{i,t}\,\log p_{\phi}\!\left(o_{i,t}\mid c_{1:m},\, o_{i,<t}\right)
}{
\sum_{t=1}^{L_i} m_{i,t}
},
\label{eq:mask_logp}
\end{equation}
where $p_{\phi}$ denotes the LLM-AD$^*$, $L_i$ is the response length, and $m_{i,t} \in \{0,1\}$ is a token-level mask indicating whether token $o_{i,t}$ is retained for scoring. 
%
% In particular, $m_{i,t}=0$ if $o_{i,t}$ belongs to a response $3$-gram or $4$-gram that also appears in the context sequence $c_{1:m}$; otherwise, $m_{i,t}=1$. 
% If all response tokens are masked out, we fall back to using all tokens for stability.

\noindent\textbf{Coherence reward.}
We then rank all eligible responses within the same rollout group according to their coherence scores $s_i$ in descending order. 
To make the reward signal more robust, we discretize the ranked scores into a small number of reward levels rather than using the raw language-model scores directly. 
Specifically, only the top-$k$ ranked responses receive a positive coherence reward, with the reward magnitude determined by their rank level, while the remaining responses receive zero additional reward. 
The resulting coherence reward is defined as
\begin{equation}
R_i^{\mathrm{coh}} =
\begin{cases}
\lambda_{\operatorname{rank}(s_i)}, &
\text{if } R_i^{\mathrm{acc}} \ge \tau_{\mathrm{acc}},\ 
\operatorname{rank}(s_i) \le k,\\
% & \quad o_i \neq \emptyset,\ \alpha_i \le \tau_{\mathrm{copy}},\\[1mm]
0, & \text{otherwise},
\end{cases}
\end{equation}
where $\lambda_{\operatorname{rank}(s_i)}$ denotes a rank-dependent discretized coherence reward, $k$ is the number of rewarded top-ranked responses.
% The concrete values used in our implementation are provided in Section~X.

\subsection{Overall READ Training}
\label{subsec:method_overll}
Our final READ framework is trained by combining the four reward components described above into a unified reward:
\begin{equation}
R_i = R_i^{\mathrm{acc}} + \lambda_{1} R_i^{\mathrm{fmt}} + \lambda_{2} R_i^{\mathrm{len}} + \lambda_{3} R_i^{\mathrm{coh}},
\end{equation}
where the accuracy reward $R_i^{\mathrm{acc}}$ serves as the anchor signal, and the remaining rewards are scaled accordingly. The resulting reward is used for policy optimization under the GRPO framework.

% =============================================================================
\section{Experiment}
\label{sec:experiment}

% We describe the experimental setup (\S\ref{subsec:ex_setting}), then compare READ with prior methods (\S\ref{subsec:ex_sota}), analyze key design choices through ablations and contextual evaluation (\S\ref{subsec:ablation}, \ref{subsec:context_eval}), and finally present the qualitative results (\ref{subsec:vis}).

\subsection{Setup}
\label{subsec:ex_setting}
% \noindent\textbf{Datasets.}
% We evaluate READ on three widely-used AD benchmarks covering both movies and TV series, namely CMD-AD~\citep{han2024autoad3}, MAD-Eval~\citep{han2023autoad}, and TV-AD~\citep{xie2024autoad0}. 
% CMD-AD is a movie-domain benchmark constructed by aligning ground-truth ADs with the Condensed Movie Dataset (CMD)~\citep{bain2020condensed}, containing $\sim$101K ADs from 1.4K movies, with 94K used for training and 7K for testing. 
% MAD-Eval is another movie-domain benchmark built on LSMDC~\citep{rohrbach2015lsmdc}, comprising about 6.5K ADs sampled from 10 movies for evaluation. 
% We further evaluate on TV-AD, a TV-series benchmark containing around 34K AD annotations from 13 series, with roughly 3K test ADs sourced from TVQA~\citep{lei2018tvqa}.
\noindent\textbf{Datasets.}
We evaluate READ on three standard datasets: CMD-AD~\citep{han2024autoad3}, MAD-Eval~\citep{han2023autoad}, and TV-AD~\citep{xie2024autoad0}. 
For CMD-AD, we train and evaluate on its official split, which contains $\sim$101K ADs from 1,432 movies, with 100 movies reserved for evaluation. 
For MAD-Eval, we follow the common protocol of training on MAD-v2-Named~\citep{soldan2022mad}, which contains $\sim$330K ADs from 488 movies, and evaluating on MAD-Eval, which consists of 6,520 ADs from 10 movies. 
For TV-AD, we use the official testing split with 3K evaluation ADs collected from 13 TV series.

\noindent\textbf{Evaluation metric.}
% Following prior evaluation protocols, 
We report conventional captioning metrics including \underline{CIDEr}~\citep{vedantam2015cider}, \underline{ROUGE-L}~\citep{lin2004rouge}, \underline{METEOR}~\citep{banerjee2005meteor}, \underline{BLEU-1}~\cite{papineni2002bleu}, \underline{SPICE}~\citep{anderson2016spice}
and AD-oriented evaluation metrics including \underline{Recall@k/N}~\citep{han2023autoad2}, \underline{Action}~\citep{xie2025shot} and \underline{LLM-AD-Eval}~\citep{han2024autoad3}. 
Detailed introductions are in \S\ref{app:eval_metric}.

% Specifically, we use  to measure lexical overlap and overall generation quality with respect to human-authored references. Following recent AD benchmarks, we further report Recall@$k/m$, which evaluates whether a generated AD can be matched to the correct reference among $m$ temporally neighboring ground-truth descriptions, thereby reflecting the distinctiveness of the output. We also report Action to assess action-level semantic correctness, and LLM-AD-Eval to measure overall AD quality with an LLM-based judge. For LLM-AD-Eval, we follow the evaluation protocol and prompt setting of prior work.

\noindent\textbf{Implementation details.}
We instantiate READ on Qwen3-VL-8B~\citep{bai2025qwen3vl} as the main backbone and fine-tune GPT-2 model~\citep{radford2019gpt-2} on pure AD corpus as LLM-AD$^*$. 
For coherence RL, we construct and select context-aware training samples \texttt{\{context,clip\}$\rightarrow$AD} according to their average temporal proximity. The ratio between these instances and standard \texttt{clip$\rightarrow$AD} pairs is set to 3:1.
% For reward design, we use the ROUGE-based accuracy reward as the anchor signal, and scale the remaining rewards accordingly. Specifically, 
Based on the true scale of accuracy reward $R^{\mathrm{acc}}$,
the weights for the $R^{\mathrm{fmt}}, R^{\mathrm{len}}, R^{\mathrm{coh}}$ are set to $0.08$, $0.05$, and $0.05$, respectively. The accuracy threshold for activating coherence is set to $\tau_{\mathrm{acc}}=0.12$. 
For rank-based coherence discretization, the top-2 responses in each rollout group receive additional coherence rewards.
Detailed experimental settings can be seen in \S\ref{app:more_ex}.

% \noindent\textbf{Implementation details.}
% Our READ framework is built upon Qwen3-VL-8B~\citep{bai2025qwen3vl} as the backbone model. To support coherence RL, we construct context-aware training instances in the form of \textit{context AD -- clip -- predicted AD}, where contextual ADs are selected based on their average temporal distance to the target clip. The resulting context-aware data are mixed with standard clip--AD pair data at a ratio of 3:1.

% In reward optimization, the accuracy reward serves as the anchor signal, and the format, length, and coherence rewards are weighted by $0.08$, $0.05$, and $0.05$, respectively. We set the accuracy gate for coherence reinforcement to $\tau_{\mathrm{acc}}=0.12$. For coherence reward discretization, the top-2 ranked responses within each rollout group are assigned additional coherence rewards.

\begin{table*}[!htbp]
  \centering
  \resizebox{\linewidth}{!}{%
  \begin{tabular}{lcccccccccc}
    \toprule
    \textbf{Method} & \textbf{VLM} & \textbf{LLM} & \textbf{Train?} & \textbf{BLEU-1} & \textbf{METEOR} & \textbf{ROUGE-L} & \textbf{SPICE} & \textbf{CIDEr} & \textbf{R@5/16} & \textbf{Action}$^{\textbf{*}}$\\
    \midrule
    % MM-Narrator~\citep{zhang2024mm} & \ding{55} & 12.8 & 6.7 & 13.4 & 5.2 & 13.9 & 49.0 & --\\
    MM-Narrator & CLIP-L14 & GPT-4 & \ding{55} & 12.8 & 6.7 & 13.4 & 5.2 & 13.9 & 49.0 & --\\
    % LLM-AD~\citep{chu2024llm-ad} & \ding{55} & -- & -- & 13.5 & -- & 20.5 & -- & --\\
    LLM-AD & GPT-4V & -- & \ding{55} & -- & -- & 13.5 & -- & 20.5 & -- & --\\
    % AutoAD-Zero~\citep{xie2024autoad0} & \ding{55} & 13.7 & 6.7 & 14.3 & 8.1 & 25.4 & 54.3  & 26.7 \\
    AutoAD-Zero & Qwen2-VL-7B & LLaMA3-8B & \ding{55} & 13.6 & 6.6 & 14.6 & 7.8 & 23.6 & 51.3  & -- \\
    AutoAD-Zero & GPT-4o & GPT-4o & \ding{55} & 13.7 & 6.7 & 14.3 & 8.1 & 25.4 & 54.3  & 26.7 \\
    % Shot-by-Shot~\citep{xie2025shot} & \ding{55} & \underline{15.9} & \underline{7.4} & 15.0 & \underline{8.5} & 26.9 & \underline{56.4} & \underline{33.5} \\
    Shot-by-Shot & Qwen2-VL-7B & LLaMA3-8B & \ding{55} & \third{16.2} & 7.2 & 14.7 & 7.8 & 25.0 & 50.6 & 27.9 \\
    Shot-by-Shot & GPT-4o & GPT-4o & \ding{55} & 15.9 & \third{7.4} & 15.0 & \third{8.5} & 26.9 & \third{56.4} & \third{33.5} \\
    \hline
    % AutoAD-I~\citep{han2023autoad} & \ding{51} & -- & -- & 11.9 & 4.4 & 13.4 & 42.1 & -- \\
    AutoAD-I & CLIP-B32 & GPT-2 & \ding{51} & -- & -- & 11.9 & 4.4 & 13.4 & 42.1 & -- \\
    % AutoAD-II~\citep{han2023autoad2} & \ding{51} & -- & -- & 13.4 & -- & 19.5 & 51.3 & -- \\
    AutoAD-II & CLIP-B32 & GPT-2 & \ding{51} & -- & -- & 13.4 & -- & 19.5 & 51.3 & -- \\
    % AutoAD-III~\citep{han2024autoad3} & \ding{51} & 13.1 & 5.5 & 13.9 & 6.1 & 24.0 & 52.8 & -- \\
    AutoAD-III & EVA-CLIP & LLaMA2-7B & \ding{51} & 13.1 & 5.5 & 13.9 & 6.1 & 24.0 & 52.8 & -- \\
    % MovieSeq~\citep{lin2024movieseq} & \ding{51} & -- & -- & 15.5 & 7.0 & 24.4 & 51.6 & -- \\
    MovieSeq & CLIP-B16 & LLaMA2-7B & \ding{51} & -- & -- & 15.5 & 7.0 & 24.4 & 51.6 & -- \\
    % DistinctAD~\citep{fang2025distinctad} & \ding{51} & -- & -- & \underline{17.6} & 8.3 & 27.3 & 56.0 & -- \\
    DistinctAD & CLIP$_{AD}$-B16 & LLaMA3-8B & \ding{51} & -- & -- & \third{17.6} & 8.3 & 27.3 & 56.0 & -- \\
    % Uni-AD~\citep{wang2025uniad} & \ding{51} & -- & -- & 17.2 & -- & \underline{28.2} & 54.9 & -- \\
    Uni-AD & CLIP-L14 & LLaMA3-8B & \ding{51} & -- & -- & 17.2 & -- & \third{28.2} & 54.9 & -- \\
    \textbf{READ (Ours)} & \multicolumn{2}{c}{Qwen2-VL-7B} & \ding{51} & \second{20.7}  & \second{9.4} & \second{20.4} & \second{9.8} & \second{39.5} & \second{60.2} & \second{35.6} \\
    \textbf{READ (Ours)} & \multicolumn{2}{c}{Qwen3-VL-8B} & \ding{51} & \first{22.7} & \first{10.4} & \first{20.8} & \first{10.9} & \first{40.0} & \first{61.7} &\first{36.1} \\
    \bottomrule
  \end{tabular}
  }
  \caption{
    \textbf{Quantitative comparisons on MAD-Eval.} 
    \colorbox{topone}{Best}, \colorbox{toptwo}{second best}, and \colorbox{topthree}{third best} denote the top-3 results for each metric.
    \textbf{*} denotes Action scores evaluated using the official Shot-by-Shot implementation~\citep{xie2025shot}.
  }
\label{tab:mad_eval}
\end{table*}

\begin{table*}[!t]
\centering
\resizebox{0.95\linewidth}{!}{%
\begin{tabular}{lcllll llll}
\toprule
\multirow{2}{*}{\textbf{Method}} & \multirow{2}{*}{\textbf{Train?}} & \multicolumn{4}{c}{\textbf{CMD-AD}} & \multicolumn{4}{c}{\textbf{TV-AD}} \\
\cmidrule(lr){3-6} \cmidrule(lr){7-10}
& & \textbf{CIDEr}  & \textbf{Action} & \textbf{R@1/5} & \textbf{LLM-AD-Eval}
& \textbf{CIDEr}  & \textbf{Action} & \textbf{R@1/5} & \textbf{LLM-AD-Eval} \\
\midrule
Video-LLaMA%~\citep{zhang2023videollama} 
& \ding{55}
& 4.8 & -- & 22.0 & 1.89\ |\ --
& -- & -- & -- & -- \\

Video-BLIP%~\citep{}
& \ding{55}
& 5.2 & -- & 23.6 & 1.91\ |\ --
& -- & -- & -- & -- \\

AutoAD-Zero%~\citep{xie2024autoad0}
& \ding{55}
& 22.4 & 30.7 & 32.9 & 3.08\ |\ \third{2.49}
& \third{30.9} & \third{26.8} & \third{34.7} & \third{3.12}\ |\ 2.57 \\

Shot-by-Shot%~\citep{xie2025shot}
& \ding{55}
& \second{26.1} & \second{32.5} & \second{36.5} & 3.17\ |\ \first{2.66}
& \second{34.2} & \second{27.4} & \second{36.6} &\second{3.12}\ |\ 2.59 \\
\hline
AutoAD-II%~\citep{han2023autoad2}
& \ding{51}
& 13.5 & -- & 26.1 & 2.08\ |\ --
& -- & -- & -- & -- \\

AutoAD-III%~\citep{han2024autoad3}
& \ding{51}
& \third{25.0} & \third{31.5} & 31.2 & 2.89\ |\ 2.01
& 26.1 & 26.4 & 30.1 & 2.78\ |\ 1.99 \\

Uni-AD%~\citep{wang2025uniad}
& \ding{51}
& 21.8 & -- & -- & 2.92\ |\ -- 
& 26.6 & -- & -- & 2.89\ |\ -- \\

DistinctAD%~\citep{fang2025distinctad}
& \ding{51}
& 22.7 & -- & \third{33.0} & 2.88\ |\ 2.03
& 27.4 & -- & 32.1 & 2.89\ |\ 2.00 \\

% no coherence
% \textbf{Ours (Qwen3-VL-8B)} & 32.4 & 32.1 & 36.5 & 3.01\ | -- & & & \\
% w/ coherence
\textbf{READ (Ours)} & \ding{51} & \first{33.7} & \first{34.9} & \first{38.0} & \second{\textbf{3.24}}\ | 2.55 & \first{40.9} & \first{30.7} & \first{40.0} & \first{3.16}\ | \first{2.62} \\
\bottomrule
\end{tabular}%
}
\caption{\textbf{Quantitative comparison on CMD-AD and TV-AD.}
\colorbox{topone}{Best}, \colorbox{toptwo}{second best}, and \colorbox{topthree}{third best} denote the top-3 results for each metric.
For both evaluations, we report results with Qwen3-VL-8B as the backbone.
}
\vspace{-4mm}
\label{tab:cmd_tvad}
\end{table*}

\subsection{Comparisons with previous methods}
\label{subsec:ex_sota}

We conduct comprehensive comparisons with prior training-free and training-based methods on widely-adopted MAD-Eval, CMD-AD and TV-AD.

\noindent\textbf{Comparisons on MAD-Eval }are shown in Tab.~\ref{tab:mad_eval}.
Very early AD systems including ClipCap~\citep{mokady2021clipcap}, ClipDec~\citep{nukrai2022clipdec}, and MM-Vid~\citep{lin2023mmvid} are omitted due to space limitation.
We group previous studies into \textit{Training-free} (upper rows) and \textit{Training-based} (bottom rows) approaches, as described in Sec.~\ref{sec:intro}.
Existing training-based methods mainly adapt frozen visual encoders (CLIP) and LLMs~\citep{touvron2023llama2,grattafiori2024llama3} through lightweight projector tuning, whereas READ performs full-parameter RL training of the underlying VLMs~\citep{wang2024qwen2vl,bai2025qwen3vl}. This allows the model to exploit its capacity more effectively for AD generation.
% But unlike former methods that partially fine-tuning only projectors on top of frozen visual encoders, \textit{e.g.}, CLIP, and LLMs, \textit{e.g.}, LLaMA-series~\citep{touvron2023llama2,grattafiori2024llama3}, our READ fully train a complete VLM, \textit{i.e.}, Qwen2-VL-7B~\citep{wang2024qwen2vl} and Qwen3-VL-8B~\citep{bai2025qwen3vl}, which significantly exploit the model's capacity to the AD domain.
%
With rule-based RL optimization, READ consistently outperforms all previous methods on every reported metric.
Using Qwen3-VL-8B as backbone, READ obtains striking gains on conventional captioning metrics including BLEU-1 (22.7 \textit{vs.} 15.9), CIDEr (40.0 \textit{vs.} 28.2) and ROUGE-L (20.8 \textit{vs.} 17.6).
% The leading result of BLEU-1 (22.7) shows that READ generates high quality ADs close the human annotations with lexical fidelity.
% READ also improves BLEU-1 from 15.9 to 22.7, METEOR from 7.4 to 10.4, ROUGE-L from 17.6 to 20.8, and SPICE from 8.5 to 10.9, demonstrating stronger lexical fidelity and overall generation quality. 
For AD-specific metrics, READ achieves the best R@5/16 (61.7 \textit{vs.} 56.4) and Action score (36.1 \textit{vs.} 33.5), indicating that the generated ADs are not only closer to human references but also more distinctive and semantically precise. 
These large and consistent improvements confirm the effectiveness of READ on AD generation.

Among training-free approaches~\citep{chu2024llm-ad,zhang2024mm,xie2024autoad0,xie2025shot}, Shot-by-Shot achieves the strongest performance through careful shot threading, handcrafted text processing, and prompting, even surpassing most prior training-based methods. 
However, compared with prompt-heavy training-free pipelines, READ consistently outperforms these works by clear margins, highlighting the stronger potential of training-based AD systems for domain adaptation and practical deployment. 
% This result suggests that, compared with prompt-heavy training-free pipelines, training-based AD systems offer greater potential for effective domain adaptation and practical deployment.

\noindent\textbf{Comparisons on CMD-AD and TV-AD.}
We further examine the generalizabilty of READ on two extra benchmarks, with results reported in Tab.~\ref{tab:cmd_tvad}. 
% For both benchmarks, we employ the Qwen3-VL-8B model.
%
Consistent with the results on MAD-Eval, READ also achieves the best overall performance on AD-oriented metrics.
On CMD-AD, READ improves previous best CIDEr, Action, and R@1/5 scores from 26.1, 32.5, and 36.5 to 33.7, 34.9, and 38.0, respectively, and also obtains the highest LLM-AD-Eval score with LLaMA2 (3.24).
READ likewise ranks first on all metrics on TV-AD, demonstrating strong generalizability from different perspectives.
One exception is the second LLM-AD-Eval score evaluated by GPT-3.5 on CMD-AD, READ performs slightly lower than Shot-by-Shot (2.55 \textit{vs.} 2.66). A possible reason is that Shot-by-Shot uses GPT-4o to generate ADs, which may enjoy an advantage when evaluated by a GPT-series judge.
Nevertheless, READ remains clearly superior on the core reference-based and semantic metrics, showing that RL with our reward design yields substantial gains in both descriptive fidelity and action-level understanding.

\noindent\textbf{Quantitative comparisons on coherence.}
Unlike CoherentAD~\citep{khandelwal2025coherentAD}, which assesses coherence indirectly through story recall and repetition-based measures, we directly quantify coherence using our tuned LLM-AD scorer (Fig.~\ref{fig:method}, bottom) via the average conditional log-probability in Eq.~\ref{eq:mask_logp}.
% The results are presented in Tab.~\ref{tab:coh}.
% Our full model, READ w/ $R^{\mathrm{coh}}$, consistently performs best, achieving the highest LogP (-3.61) and Masked LogP (-3.96), together with the lowest PPL (65.7). In contrast, removing the coherence reward leads to clear degradation on all metrics, with LogP dropping to -4.19, Masked LogP to -4.56, and PPL rising sharply to 331.6. These results provide direct evidence that the proposed coherence reinforcement substantially improves contextual compatibility between generated ADs and preceding descriptions. Moreover, READ w/ $R^{\mathrm{coh}}$ surpasses all baseline methods, including AutoAD-III, confirming the advantage of explicitly modeling coherence during training. The strong gain on Masked LogP is particularly important, as it indicates improved context-aware generation beyond simple lexical copying from the context.
As shown in Tab.~\ref{tab:coh}, the full READ model with $R^{\mathrm{coh}}$ achieves the best LogP (-3.61) and Masked LogP (-3.96), as well as the lowest PPL (99.5).
Removing the coherence reward leads to consistent degradation on all metrics, confirming that our coherence reinforcement effectively improves contextual compatibility between generated ADs and their preceding descriptions.
% indicating that our READ strategy with merging context-aware training samples and introducing coherence reward $R^{\mathrm{coh}}$ significantly improve the coherence among consecutive ADs, which makes our AD generations not only more accurate, but also more acceptable and experienced.
%
This trend also suggests that training-free methods, \textit{i.e.}, AutoAD-Zero and Shot-by-Shot, are generally weaker at maintaining narrative continuity, likely because they do not explicitly model context.

% \noindent\textbf{Backbones}
% ...

\begin{table}
  \centering
  \resizebox{\linewidth}{!}{%
  \setlength{\tabcolsep}{2pt}
  \begin{tabular}{lcccc}
    \toprule
    Method & Train? & LogP$\uparrow$ & Masked LogP$\uparrow$ & PPL$\downarrow$ \\
    \midrule
    AutoAD-Zero & \ding{55} & -4.40 & -4.57 & 273.6 \\
    Shot-by-Shot & \ding{55} & -5.21 & -5.28 & 553.0 \\
    AutoAD-III & \ding{51} & -3.91 & -4.32 & 121.0 \\
    \rowcolor{gray!10}
    \textbf{READ-Base} (w/o $R^{\mathrm{coh}}$) & \ding{51} & -4.19 & -4.56 & 222.8 \\
    \rowcolor{gray!10}
    \textbf{READ (w/ $R^{\mathrm{coh}}$)} & \ding{51} & \textbf{-3.61} & \textbf{-3.96} & \textbf{99.5} \\
    \bottomrule
  \end{tabular}
  }
  \vspace{-2.5mm}
  \caption{\textbf{Quantitative coherence evaluation on CMD-AD.}
  LogP denotes the average log-probability of generated ADs conditioned on previous ADs, Masked LogP (Eq.~\ref{eq:mask_logp}) denotes the corresponding anti-copy masked average log-probability. PPL is short for perplexity.
  }
  \label{tab:coh}
  \vspace{-2.5mm}
\end{table}

\begin{figure*}
  \includegraphics[width=1.0\linewidth]{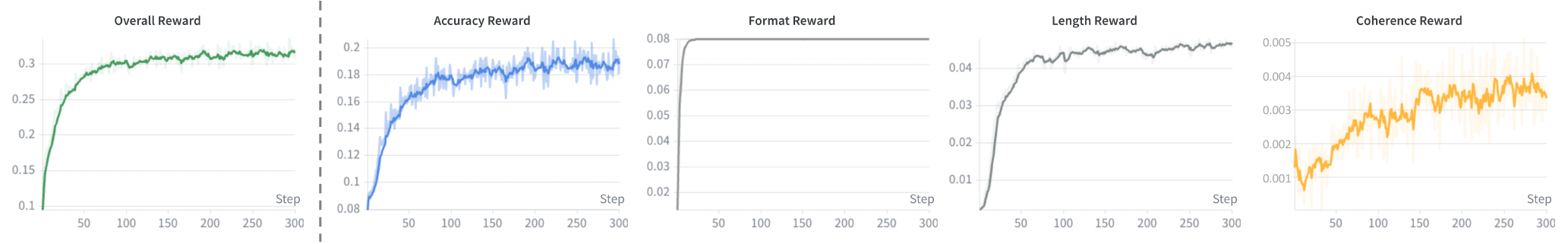}
  \vspace{-8mm}
  \caption {\textbf{Reward curves during RL training.}}
  \label{fig:train_curve}
\end{figure*}

\subsection{Ablation analysis}
\label{subsec:ablation}

\begin{table}
\centering
\resizebox{\linewidth}{!}{%
\setlength{\tabcolsep}{3pt}
\begin{tabular}{llcccc}
\toprule
\multirow{2}{*}{\#.} & \multirow{2}{*}{Setting} & \multicolumn{4}{c}{MAD-Eval} \\
\cmidrule(lr){3-6} 
& & CIDEr & BLEU-1  & R@5/16 & Action \\
\hline
A.0 & Qwen3-VL-8B (0-shot) & 4.1  & 9.7 & 55.3 & \textbf{38.0} \\
A.1 & Qwen3-VL-8B (SFT)$^{\dagger}$ & 30.8 & 15.5 & 56.5 & 33.3 \\
A.2 & Baseline ($R^{\mathrm{acc}}$) & \multicolumn{4}{c}{\textit{Non-compliant outputs}} \\
A.3 & Baseline ($R^{\mathrm{acc}},R^{\mathrm{fmt}}$) & 37.2 & \textbf{24.1} & 57.6 & 28.1 \\
A.4 & READ ($R^{\mathrm{acc}},R^{\mathrm{fmt}},R^{\mathrm{len}}$) & 38.5 & 21.6 & 61.0 & 35.4 \\
\rowcolor{gray!10}
A & \textbf{READ (w/ $R^{\mathrm{coh}}$)} & \textbf{40.0} & 22.7 & \textbf{61.7} & 36.1 \\
\hdashline
\multirow{2}{*}{\#.} & \multirow{2}{*}{Setting} & \multicolumn{4}{c}{CMD-AD} \\
\cmidrule(lr){3-6} 
& & CIDEr & Action & R@1/5 & LLM-Eval$^*$ \\
\hline
B.0 & Qwen3-VL-8B (0-shot) & 3.0 & \textbf{36.8} & 33.9 & 2.73 \\
B.1 & READ ($R^{\mathrm{acc}},R^{\mathrm{fmt}},R^{\mathrm{len}}$) & 32.4 & 32.1 & 36.5 & 3.01 \\
\rowcolor{gray!10}
B & \textbf{READ (w/ $R^{\mathrm{coh}}$)} & \textbf{33.7} & 34.9 & \textbf{38.0} & \textbf{3.24}\\
\bottomrule
\end{tabular}%
}
\vspace{-2.5mm}
\caption{\textbf{Ablation analysis.} 
$^\dagger$ denotes supervised fine-tuning with next-token prediction. 
$^*$ denotes LLM-AD-Eval with LLaMA2-7B-Chat.
}
\vspace{-4mm}
\label{tab:ablation_main}
\end{table}

\begin{table}[t]
\centering
\resizebox{\linewidth}{!}{%
\setlength{\tabcolsep}{3pt}
\begin{tabular}{lcccccc}
\toprule
\multirow{2}{*}{Method} & \multicolumn{2}{c}{Coherence} & \multicolumn{4}{c}{CMD-AD} \\
\cmidrule(lr){2-3} \cmidrule(lr){4-7} 
& \textit{gating} & \textit{masking} & CIDEr & Action & R@1/5 & LLM-Eval \\
\hline
Baseline & \ding{55} & \ding{55}  & 32.4  & 32.1 & 36.5 & 3.01 \\
\hdashline
$R^{\mathrm{coh}}$ w/ gating & \ding{51} & \ding{55} & 31.7 & 32.3 & 34.7 & 3.02 \\
$R^{\mathrm{coh}}$ w/ masking & \ding{55} & \ding{51} & 30.0 & 33.6 & 35.7 & 2.97 \\
\rowcolor{gray!10}
\textbf{READ} & \ding{51} & \ding{51} & \textbf{33.7} & \textbf{34.9} & \textbf{38.0} & \textbf{3.24}\\
\bottomrule
\end{tabular}%
}
\vspace{-3mm}
\caption{\textbf{Coherence reward design.} 
\vspace{-5mm}
}
\label{tab:coherence_setting}
\end{table}

\begin{figure*}
  \includegraphics[width=1.0\linewidth]{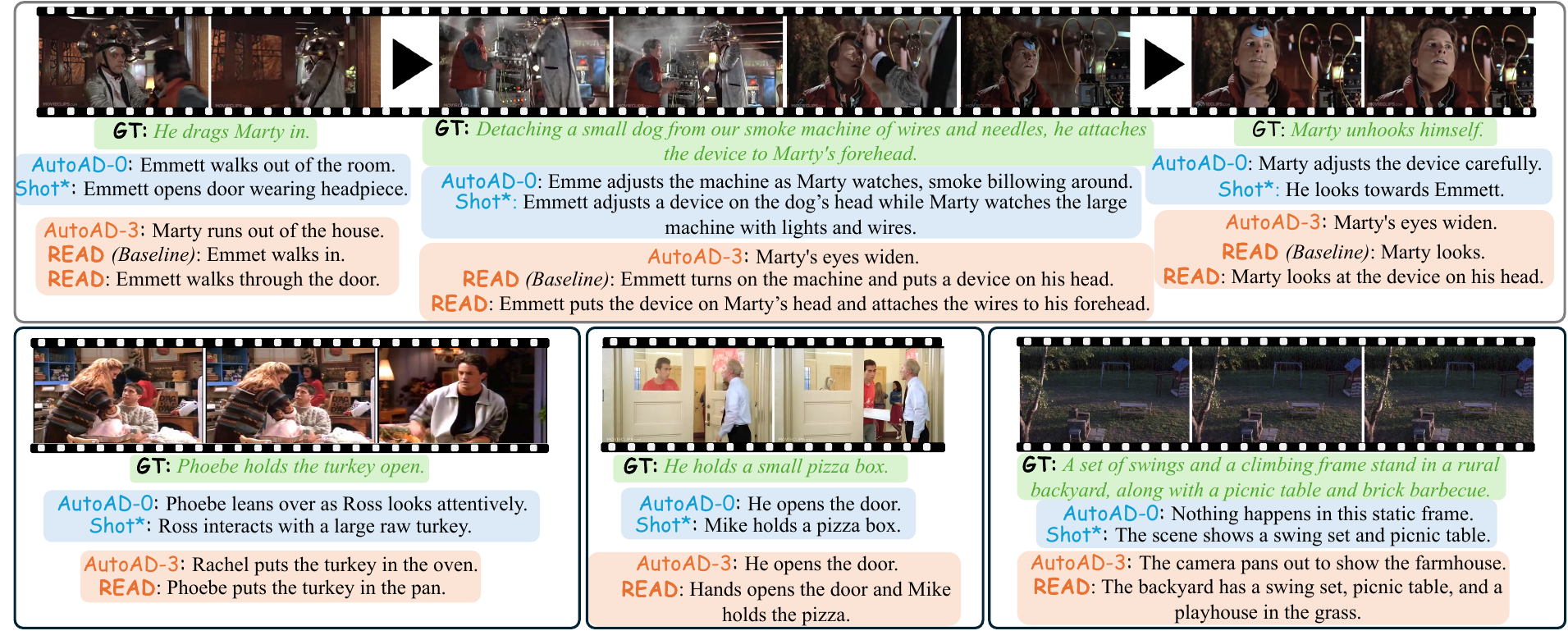}
  \vspace{-8mm}
  \caption {\textbf{Qualitative comparison of AD results.}
  GT is the ground-truth. Shot$^*$ is short for Shot-by-Shot.
  % \colorbox{myblue}{Blue-shaded boxes} denote \textit{training-free} methods, while \colorbox{toptwo}{orange-shaded boxes} denote \textit{training-based} methods.
  \colorbox{myblue}{\textit{Training-free}} and \colorbox{toptwo}{\textit{training-based}} methods are highlighted seperately. 
  Video frames are taken from \textit{Back to the Future} (1985), \textit{Friends} (S3E2), \textit{Fast Times at Ridgemont High} (1982), and \textit{Signs} (2002).
  Zoom in for details.
  }
  \vspace{-3mm}
  \label{fig:vis}
\end{figure*}

\noindent\textbf{Training mechanism.}
We first verify the effectiveness of RL for AD generation (Tab.~\ref{tab:ablation_main}). 
% On MAD-Eval, while SFT substantially improves over the zero-shot backbone (A.1 \textit{vs.} A.0), RL-based READ (A) yields much larger gains, improving CIDEr from 30.8 to 40.0, BLEU-1 from 15.5 to 22.7, and R@5/16 from 56.5 to 61.7. A similar trend is observed on CMD-AD, where the zero-shot model (B.0) performs poorly on reference-based metrics, whereas RL baseline training (B.1) and the full READ model (B) improve CIDEr from 3.0 to 32.4 and further to 33.7.
On both benchmarks, RL-based READ markedly outperforms the zero-shot and SFT baselines, with the most notable gains on CIDEr, \textit{e.g.}, from 30.8 to 40.0 over SFT (Tab.~\ref{tab:ablation_main} row A.1 \textit{vs}. A), and from 3.0 to 33.7 over zero-shot (B.0 \textit{vs}. B).
These results show that RL provides a much more effective training mechanism than direct prompting or standard next-token supervised fine-tuning.
Although the zero-shot model achieves high Action scores, its outputs are overly detailed and lengthy, which increases the chance of matching GT verbs but tends to dilute salient information under other metrics; 
such behavior is also undesirable in practice, since ADs must fit within limited narration window between dialogues.

\noindent\textbf{Baseline rewards.}
Next, we study the effect of major rewards in READ, on top of the anchor $R^{\mathrm{acc}}$. 
Without format $R^{\mathrm{fmt}}$ (A.2), the model often fails to follow the required template, making it difficult to evaluate consistently. 
In A.3, removing length $R^{\mathrm{len}}$ encourages \textit{reward hacking}, with the model favoring short and generic descriptions.
While such outputs reduce the risk of obvious errors, they are also less informative, which is evidenced by the lowest Action (28.1) among all variants.
%
% After introducing the length reward (A.4), the model produces more appropriate ADs and achieves clear improvements on all metrics. 
% This confirms that the proposed reward design is both necessary and complementary.
Together, these three reward components establish a robust and effective READ baseline (A.4 \& B.1).

\noindent\textbf{Effectiveness of $R^{\mathrm{coh}}$.}
We finally evaluate the effect of the coherence reward by comparing A.4 \textit{vs.} A on MAD-Eval and B.1 \textit{vs.} B on CMD-AD in Tab.~\ref{tab:ablation_main}.
Despite $R^{\mathrm{coh}}$ is mainly designed to improve coherence with preceding ADs, it also consistently yields descent and stable gains on standard AD quality metrics. 
This indicates that, under the accuracy-gating design, coherence RL does not compromise descriptive fidelity, but instead helps the model refine its outputs into descriptions that are context-compatible and narratively coherent.

\noindent\textbf{Coherence reward mechanism.}
\textit{Accuracy gating} and \textit{anti-copy masking} are important in coherence reward assignment (Tab.~\ref{tab:coherence_setting}).
The gating mechanism restricts coherence reinforcement to responses  sufficiently faithful to the current clip, making coherence optimization more controlled. 
% removing it leads to clear degradation on reference-based quality metrics such as CIDEr. 
Anti-copy masking prevents the model from obtaining coherence reward by simply repeating contexts. 
Without masking, R@1/5 drops markedly (34.7), indicating reduced distinctiveness and clear reward hacking.

\subsection{Training curves}
\label{app:curves}

Fig.~\ref{fig:train_curve} presents the reward curves during RL training. 
All curves show stable and reasonable trends, indicating that the optimization process is well-behaved. 
The format and length rewards increase rapidly and quickly plateau, suggesting that the model soon learns to follow the required output format and control the description length. 
The accuracy reward rises more gradually and saturates around 0.18, reflecting steady improvement in reference alignment. 
Although the coherence reward remains much smaller in magnitude due to the accuracy-gating mechanism and top-$k$ reward assignment, it also shows a clear upward trend, indicating progressively better contextual coherence.

\subsection{Evaluation under contextual settings}
\label{subsec:context_eval}

\begin{table}
\centering
\resizebox{\linewidth}{!}{%
\setlength{\tabcolsep}{4pt}
\begin{tabular}{llccccc}
\toprule
\#. & Context & w/ $R^{\mathrm{coh}}$? & CIDEr & Action & R@1/5 & LLM-Eval$^*$ \\
\hline
\rowcolor{gray!10}
A & NA & \ding{55} & 32.4 & 32.1 & \textbf{36.5} & 3.01 \\
A.1 & Reccurrent &\ding{55} & 32.2\reddown & 31.3\reddown & 35.5\reddown & 2.99\reddown \\
A.2 & \textcolor{gray}{Oracle} & \textcolor{gray}{\ding{55}}& \textcolor{gray}{\textbf{39.7}} & \textcolor{gray}{\textbf{32.4}} & \textcolor{gray}{34.9} & \textcolor{gray}{\textbf{3.07}}\\
\hdashline
\rowcolor{gray!10}
B & NA & \ding{51} & 33.7 & 34.9 & \textbf{38.0} & 3.24 \\
B.1 & Reccurrent &\ding{51} & 34.2\greenup & 35.2\greenup & 36.1\reddown & 3.25\greenup \\
B.2 & \textcolor{gray}{Oracle} & \textcolor{gray}{\ding{51}} & \textcolor{gray}{\textbf{39.6}} & \textcolor{gray}{\textbf{36.4}} & \textcolor{gray}{34.2} & \textcolor{gray}{\textbf{3.30}}\\
\bottomrule
\end{tabular}%
}
\caption{\textbf{Context evaluation on CMD-AD.} 
Arrows indicate whether performance improves or degrades relative to the recurrent setting.
$^*$ denotes LLM-AD-Eval with LLaMA2-7B-Chat.
}
\vspace{-4mm}
\label{tab:ctx_eval}
\end{table}

Since READ is trained with extra context-aware samples, it can naturally operate in a \textit{recurrent} evaluation setting, where previous AD predictions are used as contexts~\citep{han2023autoad}. 
As shown in Tab.~\ref{tab:ctx_eval}, coherence modeling is crucial in this setting: w/ $R^{\mathrm{coh}}$, recurrent evaluation brings additional gains (B.1 \textit{vs.} B), whereas w/o $R^{\mathrm{coh}}$, introducing context instead degrades performance. 
% This suggests that context is beneficial only when the model is explicitly trained to use it coherently.
With GT contexts (\textit{oracle}), CIDEr improves substantially, whereas R@1/5 drops, indicating reduced distinctiveness. 
This suggests that the model may over-attend on the context and reproduce contextual content, rather than generating sufficiently discriminative ADs for the current clip.

% \subsection{Training curves}
% \label{subsec:curves}

% Fig.~\ref{fig:train_curve} presents the reward curves during RL training. 
% All curves show stable and reasonable trends, indicating that the optimization process is well-behaved. 
% %
% The format and length rewards increase rapidly and quickly plateau, suggesting that the model soon learns to follow the required output format and control the description length. 
% %
% The accuracy reward rises more gradually and saturates around 0.18, reflecting steady improvement in reference alignment. 
% %
% Although the coherence reward remains much smaller in magnitude due to the accuracy-gating mechanism and top-$k$ reward assignment, it also shows a clear upward trend, indicating progressively better contextual coherence.

\subsection{Qualitative results}
\label{subsec:vis}
Fig.~\ref{fig:vis} shows qualitative comparisons of AD generation. 
In the top row, consisting of three \textit{consecutive} clips, READ baseline (w/o $R^{\mathrm{coh}}$) is more visually accurate than prior methods, while the full READ produces more coherent cross-clip descriptions. For instance, in the last clip, READ explicitly connects the current output to the previously mentioned “device,” demonstrating better contextual continuity. 
In the bottom row, which contains three independent clips, READ generates ADs that are consistently closer to GT, further confirming its superiority. More examples are in \S\ref{app:more_vis}.

\section{Conclusion}
\label{sec:conclusion}
We presented READ, an effective RL framework for automatic AD generation. 
READ introduces RL into training-based AD generation and further improves narrative coherence through a dedicated coherence reward with accuracy gating and anti-copy masking. 
Extensive experiments on multiple AD benchmarks show that READ consistently outperforms prior training-free and training-based methods across a wide range of metrics. 
These results demonstrate that reinforcement learning provides an effective training paradigm for generating accurate, distinctive, and context-compatible ADs.

\newpage
% ****** Important ******
\section*{Limitations}
The limitations of READ includes:
First, coherence modeling in READ relies on an external LLM-AD scorer and heuristic designs such as accuracy gating, reward discretization, and anti-copy masking, which may not fully capture higher-level narrative coherence. 
In addition, the current framework mainly uses local textual context from neighboring ADs, without explicitly modeling longer-range story structure, dialogue semantics, or audio cues. 
Finally, our experiments are conducted on existing English AD benchmarks, and the generalization of READ to broader domains, longer contexts, and other languages remains to be explored.

% \section*{Acknowledgments}


\bibliography{custom}

\newpage
\appendix
\section*{Appendix}
\label{sec:appendix}
The appendix is organized as follows:
\begin{itemize}
    \item \textbf{Appendix A} reports the LLM usage statement.
    \item \textbf{Appendix B} provides experimental details, including \textit{(B.1)} the training prompts, \textit{(B.2)} the training details and hyperparameter settings for READ and the LLM-AD scorer, and \textit{(B.3)} the definitions of the evaluation metrics.
    \item \textbf{Appendix C} presents additional analyses, including \textit{(C.1)} AD length statistics, \textit{(C.2)} a comparison between ROUGE and CIDEr as the accuracy reward, and \textit{(C.3)} the effect of longer RL training.
    \item \textbf{Appendix D} provides additional qualitative visualizations.
\end{itemize}

\section{LLM Usage}
\textit{LLMs were primarily used to polish the writing of this paper, including improving fluency, grammar, and presentation.
We also used LLMs in a limited scope to assist with minor implementation refinement, mainly for code optimization related to parallel evaluation and inference efficiency. 
\textbf{GPT-5.4} was the primary LLM used. 
The core research ideas, method design, experimental setup, and final conclusions were developed and verified by the authors.}

\section{Experimental Details}
\label{app:more_ex}

\subsection{Training prompts}
\label{app:prompt}

\begin{lstlisting}[caption={Prompt of \{clip -> AD\}}, label={lst:pure_instance}]
<image><image><image><image><image><image><image><image>\n
Audio Descriptions (ADs) are written narrations that describe the visual elements of a video, including characters, actions, interactions, and scene settings, especially those not conveyed through dialogue. The video clip is 2.70 seconds long. Consider the clip duration when completing the following task. \n
Please provide a concise audio description summarizing what happens in the video clip in one sentence, focusing on key characters, actions, interactions, or important visual details.\n
Possible character portraits and their names are: <image> Jeff Kohlver.
\end{lstlisting}

READ uses two types of training prompts, as shown in Lst.~\ref{lst:pure_instance} and \ref{lst:context_instance}. The first is a standard \texttt{clip}$\rightarrow$\texttt{AD} prompt, while the second is a context-aware prompt for \texttt{\{context,clip\}$\rightarrow$AD} training.

\begin{lstlisting}[caption={Prompt of \{context, clip -> AD\}}, label={lst:context_instance}]
<image><image><image><image><image><image><image><image>\n
Audio Descriptions (ADs) are written narrations that describe the visual elements of a video, including characters, actions, interactions, and scene settings, especially those not conveyed through dialogue.\n\n

You are given a video clip with its prior and chronologically ordered contexts.\n
Context:\n
[AD] The sprinkler system goes on.\n
[AD] He drags the ringleader to his feet.\n
[AD] Ghost Rider puts on the skinheads leather jackets.\n\n

Task:\n
Please provide a concise audio description summarizing what happens in the target video clip in one sentence. The video clip is 2.75 seconds long. Consider the clip duration when summarizing ADs.
\end{lstlisting}

\subsection{Training details}
\label{app:train_detail}

\begin{table}
\centering
\resizebox{\linewidth}{!}{%
\begin{tabular}{lc}
\toprule
\textbf{Hyperparameter} & \textbf{Value} \\
\midrule
Algorithm & GRPO \\
Base model & Qwen3-VL-8B-Instruct \\
Training batch size & 32 \\
Rollout batch size & 128 \\
Rollout per prompt (n) & 8 \\
Learning rate & $2\times10^{-6}$ \\
Max prompt length & 16,384 \\
Max response length & 4,096 \\
Precision & bf16 \\
Optimizer & AdamW bf16 \\
% Tensor parallel size & 4 \\
% GPUs & 8 \\
Epochs & 1 \\
% Online filtering & Accuracy, 0.01--0.99 \\
KL coefficient & $1\times10^{-2}$ \\
\bottomrule
\end{tabular}
}
\caption{\textbf{Key hyperparameters for READ training with the VeRL framework.}}
\label{tab:verl_hyperparams}
\end{table}

Tab.~\ref{tab:verl_hyperparams} summarizes the main hyperparameter settings for fine-tuning READ on Qwen3-VL-8B-Instruct with the VeRL framework~\cite{sheng2025hybridflow}. The Qwen2-VL-7B-Instruct variant uses the same training configuration. All experiments are conducted on 8 NVIDIA A800 GPUs with 80GB memory each.

\begin{table}
\centering
\begin{tabular}{lc}
\toprule
\textbf{Hyperparameter} & \textbf{Value} \\
\midrule
Base model & GPT-2 \\
% Objective & Causal LM fine-tuning \\
Max sequence length & 1,024 \\
Batch size per device & 8 \\
Gradient accumulation & 8 \\
% Effective batch size & 16 per GPU \\
Learning rate & $1\times 10^{-4}$ \\
Epochs & 1 \\
Warmup ratio & 0.03 \\
Weight decay & 0 \\
% Precision & FP32 \\
% Seed & 42 \\
% Loss tokens & Response tokens only \\
% Save interval & 1,000 steps \\
\bottomrule
\end{tabular}
\caption{\textbf{Hyperparameters for GPT-2 fine-tuning on the AudioVault corpus.}}
\label{tab:gpt2_ad_hyperparams}
\end{table}

Tab.~\ref{tab:gpt2_ad_hyperparams} reports the hyperparameter settings for fine-tuning the LLM scorer on AudioVault\footnote{https://audiovault.net}. We use GPT-2 as the base model and train it with casual language model modeling: given the preceding three adjacent ADs as context, the model predicts the next AD. The resulting scorer is then used for coherence evaluation and reward construction in READ.

\subsection{Evaluation metrics details}
\label{app:eval_metric}
\paragraph{CIDEr}
measures the similarity between a generated description and reference descriptions using TF-IDF~\citep{salton1988term} weighted $n$-gram matching. It emphasizes informative content words and phrases.

\paragraph{ROUGE-L}
is based on the longest common subsequence (LCS) between the generated description and the reference. It reflects sequence-level overlap in content and word order.

\paragraph{METEOR}
evaluates word-level alignment between the generated description and the reference using a harmonic mean of precision and recall, with additional flexibility such as stemming and synonym matching. It is designed to better capture semantic similarity than exact lexical overlap.

\paragraph{BLEU-1}
measures unigram precision between the generated description and the reference, together with a brevity penalty. It mainly reflects local lexical overlap.

\paragraph{SPICE}
compares the semantic content of the generated description and the reference by converting them into scene-graph tuples, such as objects, attributes, and relations. It focuses on semantic agreement rather than surface-level wording.

\paragraph{Recall@k/N}
is a retrieval-based metric that evaluates whether a generated description can be matched to the correct GT AD among $N$ temporally adjacent references. For each generated AD, similarity scores with the neighboring GT ADs are computed using BERTScore~\citep{zhang2019bertscore}, and Recall@k is then measured within this local temporal window.

\paragraph{Action}
measures whether a generated description correctly captures the actions in the reference AD. It combines two components: a semantic similarity score computed with text embeddings, and a verb-matching score based on whether the predicted and GT descriptions contain the same action verbs. The final score is obtained by combining these two signals.
% , so that both overall action semantics and explicit verb matching are taken into account.

\paragraph{LLM-AD-Eval}
uses a LLM as a judge to assess the quality of a generated AD with respect to the GT AD. Following prior work, the LLM is prompted to score each prediction-reference pair on a scale from 1 to 5, where higher scores indicate better matching quality. 
% This metric is intended to capture overall AD quality beyond surface-level lexical overlap.

\paragraph{LogP}
denotes the average log-probability of the generated AD conditioned on the contextual ADs. It measures how likely the generated description is under the context according to the LLM-AD scorer.

\paragraph{Masked LogP}
refers to the corresponding masked average log-probability, where tokens that overlap with context-copying spans are excluded from the computation. It is designed to better reflect genuine contextual compatibility rather than superficial repetition.

\paragraph{PPL (perplexity)}
is computed from the average token-level log-probability and reflects the overall uncertainty of the scorer on the generated AD. Lower PPL indicates that the generated description is more probable under the given context.

% [!htbp]
\begin{table*}
  \centering
  \resizebox{\linewidth}{!}{%
  \begin{tabular}{llccccccccc}
    \toprule
    \textbf{\#.} & \textbf{Setting} & $R^{\mathrm{acc}}$ \textbf{Metric} & \textbf{BLEU-1} & \textbf{METEOR} & \textbf{ROUGE-L} & \textbf{SPICE} & \textbf{CIDEr} & \textbf{Action} & \textbf{R@1/5} & \textbf{LLM-AD-Eval}\\
    \midrule
    \rowcolor{gray!10}
    A.1 & Baseline (w/o $R^{\mathrm{coh}}$) & \textbf{ROUGE} (Eq.~\ref{eq:ROUGE}) & 18.4 & 7.2 & 16.5 & 7.5 & 32.4 & \textbf{32.1} & 36.5 & 3.01 \\
    A.2 & Baseline (w/o $R^{\mathrm{coh}}$) & \textbf{CIDEr} & \textbf{19.9} & \textbf{7.6} & \textbf{17.0} & \textbf{7.6} & \textbf{35.1} & 31.9 & \textbf{37.4} & \textbf{3.05} \\
    \hdashline
    \rowcolor{gray!10}
    B.1 & READ (w/ $R^{\mathrm{coh}}$) & \textbf{ROUGE} (Eq.~\ref{eq:ROUGE}) & 19.7 & \textbf{8.2} & 17.5 & 7.7 & 33.7 & \textbf{34.9} & \textbf{38.0} & \textbf{3.24} \\
    B.2 & READ (w/ $R^{\mathrm{coh}}$) & \textbf{CIDEr} & \textbf{20.9} & 7.9 & 17.5 & 7.7 & \textbf{35.8} & 33.1 & 37.7 & 3.17 \\
    \bottomrule
  \end{tabular}
  }
  \caption{
    \textbf{Effect of accuracy reward formulation on CMD-AD: ROUGE \textit{vs.} CIDEr.} 
  }
\label{tab:cider_vs_ROUGE}
\end{table*}

\section{More Analysis.}

\subsection{AD Length Statistics.}
\begin{figure}
  \includegraphics[width=1.0\linewidth]{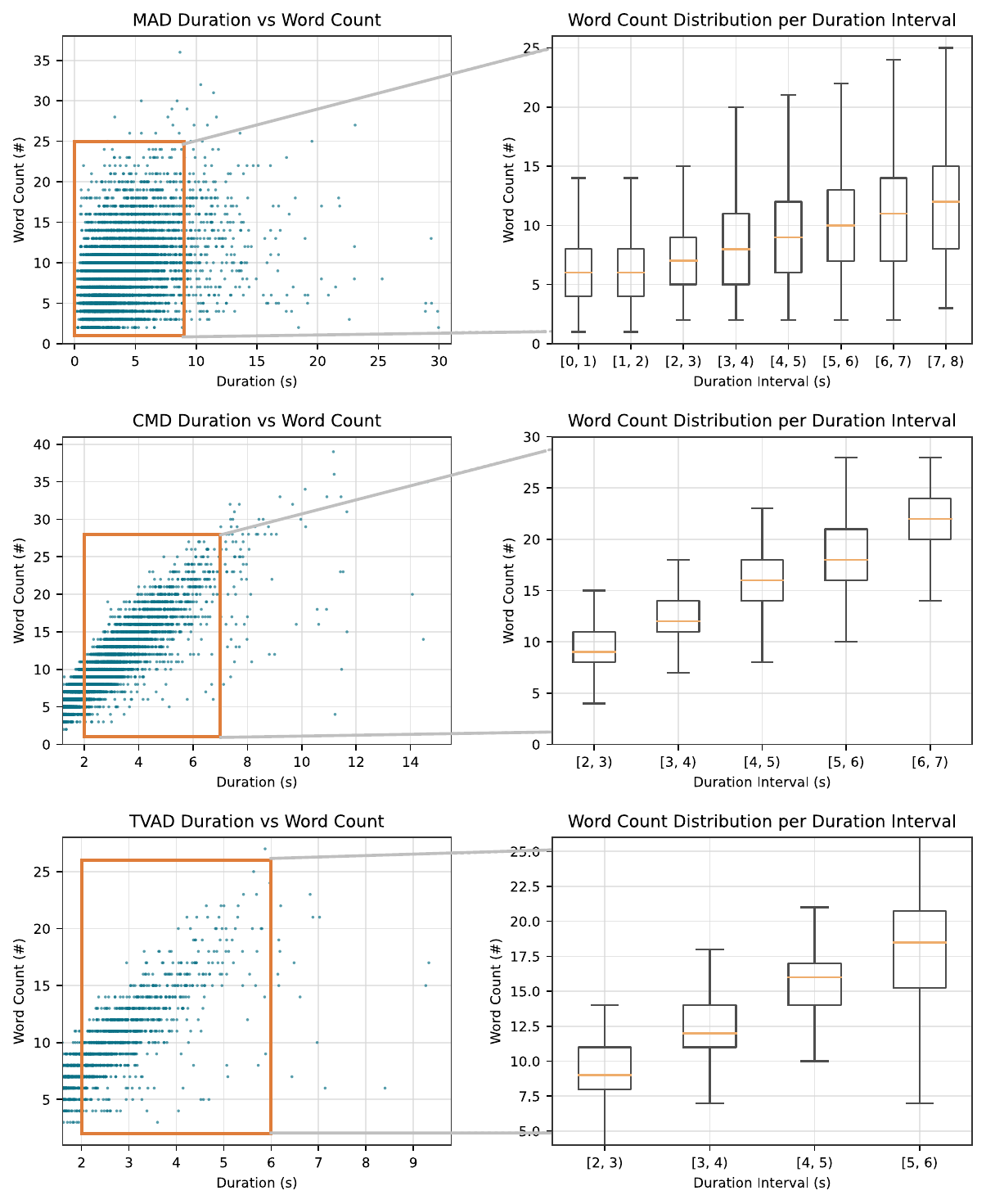}
  \caption{\textbf{Clip duration \textit{vs.} AD word count statistics.}
  }
  \label{fig:word_count}
\end{figure}
\label{app:ad_len_sta}
Fig.~\ref{fig:word_count} summarizes the statistics of AD word count and the corresponding clip duration on the three benchmarks. Two observations are particularly important. 

First, most ADs are concentrated in the lower-left region of the scatter plots, indicating that short descriptions dominate the data distribution, while longer ADs form a relatively sparse long tail. This imbalance suggests that pure next-token prediction can easily bias the model toward short and generic outputs, making RL-based optimization more suitable for controlling generation behavior. 

Second, AD length is positively correlated with clip duration across all three datasets, showing that appropriate description length should depend on the temporal span of the target clip. This motivates the introduction of an explicit length reward in READ. Since the exact duration--length relationship varies across datasets, we do not fit a dataset-specific function; instead, we design the length reward based on the distance between the generated word count and the ground-truth AD length. In practice, the acceptable range is controlled by a tolerance threshold $\Lambda=4$.

\subsection{ROUGE vs. CIDEr as the Accuracy Reward}
We further compare two choices for the anchor accuracy reward $R^{\mathrm{acc}}$, namely the average ROUGE score in Eq.~\ref{eq:ROUGE} and CIDEr. As shown in Tab.~\ref{tab:cider_vs_ROUGE}, using CIDEr as the optimization target yields slightly better performance on the final CIDEr metric itself, which is expected since the reward is more directly aligned with this evaluation measure. However, when coherence reinforcement is enabled, the ROUGE-based formulation produces consistently stronger AD-specific results, including higher Action, R@1/5, and LLM-AD-Eval scores.

A plausible explanation is that CIDEr and ROUGE provide different optimization signals. CIDEr is based on TF-IDF weighted $n$-gram matching and therefore more strongly favors exact overlap with informative reference phrases, which naturally benefits the final CIDEr metric. In contrast, the average ROUGE reward provides a smoother and more balanced sequence-level supervision signal, focusing more on overall content coverage and structural overlap. This makes it a more robust anchor reward when combined with coherence reinforcement: the model can rewrite its outputs to better fit the context without being overly penalized for moderate lexical variation. As a result, although CIDEr-based optimization yields slightly better CIDEr scores, the ROUGE-based formulation leads to stronger overall AD-specific performance.

\begin{figure}
  \includegraphics[width=1.0\linewidth]{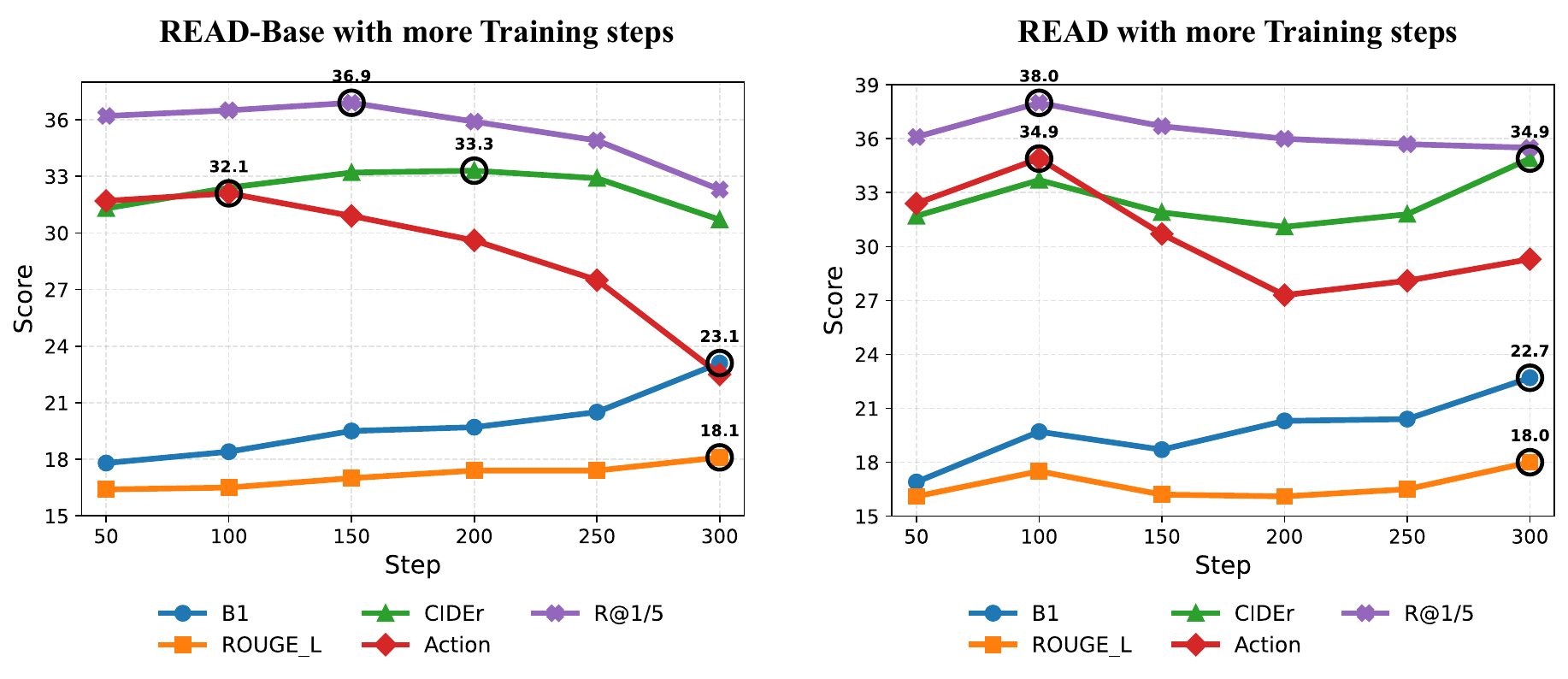}
  \caption{\textbf{Effect of longer RL training on READ-Base (left) and READ (right).}
  }
  \label{fig:more_train}
\end{figure}

\subsection{Effect of Longer RL Training}
\label{app:longer_train}
Fig.~\ref{fig:more_train} shows the effect of continuing RL optimization with more training steps for both READ-Base and READ.
We observe that longer training further improves conventional captioning metrics such as BLEU-1, ROUGE-L, and CIDEr. However, this gain does not necessarily reflect better AD quality. Instead, it suggests a form of \textit{reward hacking}: the model may increasingly rely on frequent words and generic expressions that are more likely to match the reference and thus obtain higher accuracy rewards. 
In contrast, AD-specific metrics such as Action and R@1/5 show a clear downward trend as training continues, indicating that the generated descriptions become less action-accurate and less distinctive. This discrepancy further highlights a limitation of using overlap-based metrics such as ROUGE or CIDEr as the anchor accuracy reward. How to design a more robust accuracy reward remains an important direction for future work.

\section{More Qualitative Results}
\label{app:more_vis}
\begin{figure*}
  \includegraphics[width=1.0\linewidth]{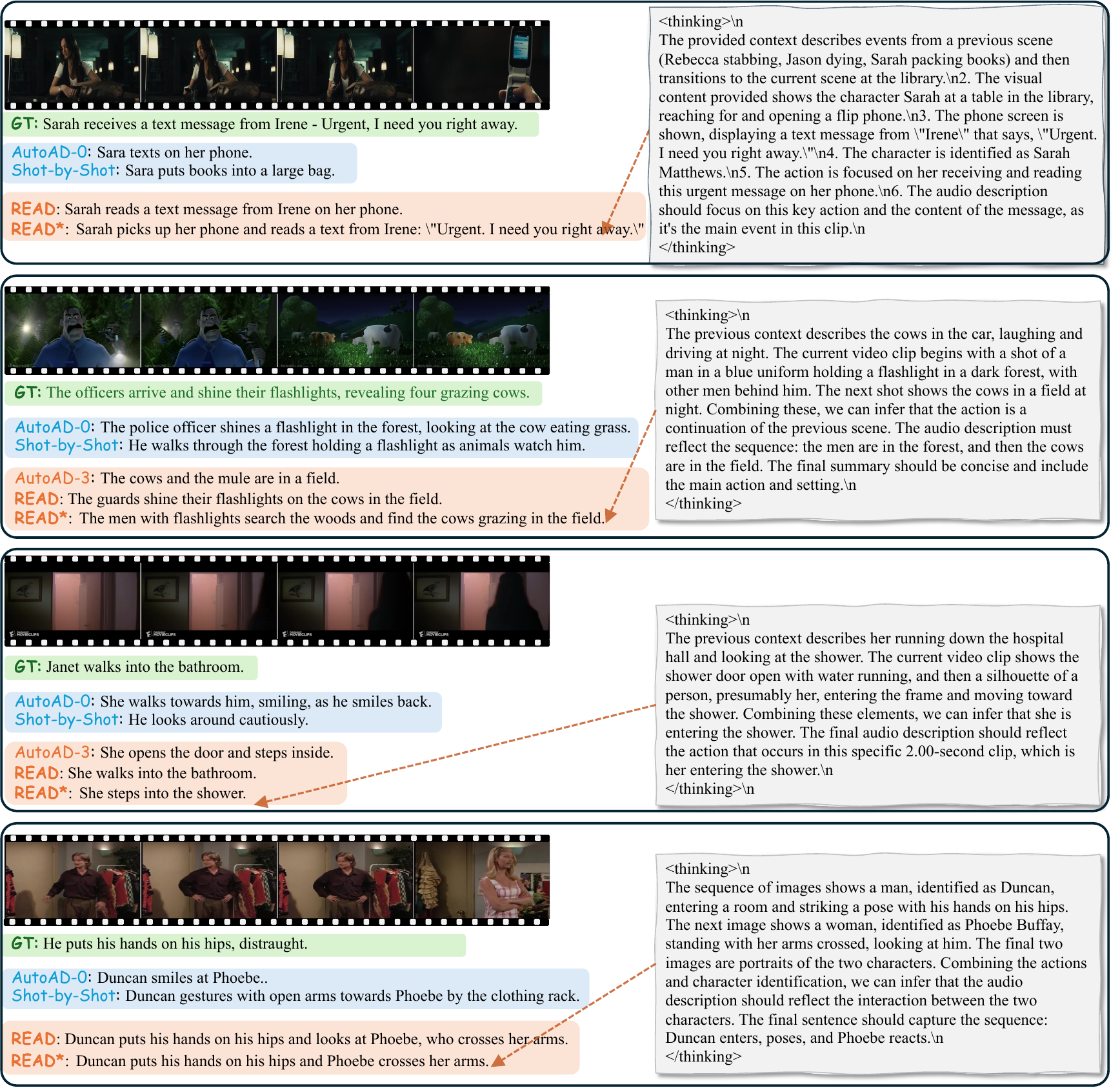}
  \caption{\textbf{Additional qualitative results.}
  READ$^{*}$ denotes recurrent inference using the previous three predicted ADs as context. 
  The right side visualizes the intermediate thinking process used to infer the final description.
  }
  \label{fig:more_vis_single}
\end{figure*}

We provide more qualitative results in Fig.~\ref{fig:more_vis_single}.
Compared with prior methods and the non-recurrent READ output, READ$^{*}$ better leverages narrative context to produce more coherent and context-compatible descriptions. 
The right side shows the intermediate thinking process, illustrating how READ reasons over (preceding AD context and) current visual content before generating the final description.

\end{document}